\documentclass[11pt]{article}
\usepackage{amsthm,amsmath,amssymb,anysize}
\usepackage{graphicx}

\def\qed{\hbox to 0pt{}\hfill$\rlap{$\sqcap$}\sqcup$}

\usepackage[numbers,sort&compress]{natbib}
 \numberwithin{equation}{subsection}

\begin{document}

\title{\textbf{Concept and Attribute Reduction Based on Rectangle Theory of Formal Concept
\footnote{The
authors are supported by NSF grant of
 Anhui Province(No.1808085MF178), China.} }}
\author{Jianqin Zhou$^{1}$ , Sichun Yang$^{1}$,  Xifeng Wang$^{1}$ and Wanquan Liu$^{2}$ \footnote{Corresponding author.  Email: liuwq63@mail.sysu.edu.cn}\\
\\
\textit{$^{1}$Department of Computer Science,} \\
\textit{Anhui University of   Technology, Ma'anshan 243002,   China}\\
\\
\textit{$^{2}$School of Intelligent Systems Engineering,} \\
\textit{Sun Yat-sen University, Shenzhen 518000,  China}}
% w.liu@curtin.edu.au
\date{ }
\maketitle
\begin{quotation}
\small\noindent

Based on rectangle theory of formal concept and set covering theory, the concept reduction preserving binary
relations is investigated in this paper. It is known that there are three types of formal concepts:  core concepts, relative necessary concepts and
unnecessary concepts. First, we present the new judgment results for relative necessary concepts and
unnecessary concepts. Second, we  derive the bounds for both the maximum number of relative necessary concepts and the maximum number of unnecessary concepts and it is a difficult problem as either in concept reduction preserving binary relations or attribute reduction of decision formal contexts, the computation of formal contexts from formal concepts is a challenging problem. Third, based on rectangle theory of formal concept, a fast algorithm for reducing attributes while preserving the extensions for a set of formal concepts is proposed
using the extension bit-array technique, which allows multiple context cells to be processed by a single 32-bit or 64-bit operator.
Technically, the new algorithm could store both formal context and extent of a concept as bit-arrays, and we can use bit-operations to process set operations "or" as well as "and". One more merit is that the new algorithm does not need to consider other concepts in the concept lattice, thus the algorithm is explicit to understand and fast.  Experiments demonstrate that the new algorithm is effective in the computation of attribute reductions.



\end{quotation}

 \section{Introduction}

Concept lattice is a key  tool for information analysis and processing. The mathematical basis of concept lattice is the lattice theory, the visualization tool is the Hasse graph, and the research methods are abstract algebra, discrete mathematics, data structure and algorithm analysis, fuzzy set \cite{Zadeh}, rough set \cite{Pawlak}, granular computing \cite{Zadeh97}, etc.
So far, formal concept analysis has been widely used in information retrieval \cite{Li,Carpineto}, knowledge discovery \cite{Nguyen}, association analysis \cite{Tu}, recommendation system \cite{Zou} and other fields \cite{PIV,PBT, SYW,WW}.

When we decide to study a certain kind of concepts, we need to first consider how to find out all the concepts from the specific data. This problem is called concept lattice construction \cite{Andrews17, Outrata,  Andrews,   Osicka,  QLW, QQW}.
Second, in order to better analyze data  and save storage space, it is necessary to reduce concept lattice \cite{Cao, LMWZ, Kuznetsov2007,Ren, Wei,Zhang}.
Furthermore, the nodes of concept lattice can infer from each other, and based on this, one can extract rules \cite{Li, LML,LMW,LMWZ,PRM}.

Based factorization and attribute reduction, the concept reduction while
preserving binary relations is proposed in \cite{Cao,Wei}. The significance of a concept in a given formal context may be different and they play different roles in
the concept reduction. These concepts can be classified into three types: the core concepts, the relatively necessary concepts, and the  unnecessary
concepts.  The judgment theorems for three types of concepts have been investigated. However, the judgment theorems for relative necessary concepts and
unnecessary concepts in \cite{Cao, Wei} are very complicated, and difficult to implemented.

One can visualise a formal concept in a  table of $0$ and $1$  as a closed rectangle of $1$s, where the rows and columns  are not necessarily contiguous \cite{Andrews}.
Based on rectangle theory of formal concept and set covering theory, we will present the new judgment theorems for relative necessary concepts and
unnecessary concepts in this paper. Also, we  deduce the bounds for both the maximum number of relative necessary concepts and the maximum number of unnecessary concepts.
From our results, it is known that both relatively necessary concepts and unnecessary concepts are exponential time problems. However,
by using the proposed judgment theorems, it is easier to develop a program for computing both  relative necessary concepts and
unnecessary concepts than that of \cite{ Wei}.

Based on the discernibility matrix and Boolean function, a method to attribute reduction is proposed while preserving the extensions
in concept lattices \cite{Zhang, Zhang05}.  Similarly,  two kinds of attribute reduction are proposed in the decision formal context based on the maximal rules \cite{Li}.
Note that the attribute reduction in both \cite{Zhang, Zhang05} and \cite{Li}  can be simplified to reducing attributes while preserving the extensions for a set of formal concepts. A fast algorithm is proposed in this paper for  reducing attributes while preserving the extensions for a set of  formal concepts.
As the new algorithm store both formal context and extension of a concept as a bit-array, and we can use bit-operations to process set operation "or" as well as "and". Further more, there is no need to consider other formal concepts, thus the algorithm is very fast.

Technically, suppose that one column of context or the extension of a concept is
$(1,1,1,1$, $1,1,1,1$, $0,0,0,0,0,0,0,0,$  $0,0,0,0,$  $0,0,0,0$,  $0,0,0,0$,   $0,0,0,0)$,  it is stored as $255$ in our algorithm, where the first bit means $2^0$, the second bit means $2^1$, the third bit means $2^2$ and so on. Applying our new  algorithm to mushroom data from \cite{Frank}, 19 columns of formal context  can be reduced while preserving the extensions for 512 formal concepts. Experiments demonstrate that the new algorithm is much efficient in the computation of attribute reductions.

The remainder of this paper is organized as follows. In Section \ref{s1}, some basic definitions and lemmas are reviewed. In Section \ref{s2}, concept reduction preserving binary relations is discussed, we present the new judgment theorems for relative necessary concepts and
unnecessary concepts. In Section \ref{s03}, based on rectangle theory of formal concept, a fast algorithm for  reducing attributes while preserving the extensions for a set of formal concepts is proposed by using extension bit-array technique.
Finally, in Section  \ref{s5}, we conclude the paper with a summary and an outlook for  future work.

\section{Basic notions and properties} \label{s1}

%\subsection{Basic notions and properties}\label{sub1.1}
For the convenience of discussion,  some basic notions and properties of formal concepts related to this paper will be reviewed.
 We first present the formal context and its operators as follows.

\textbf{Definition 1}. \cite{Ganter} A triplet  $(U, A, I)$ is called a formal context, where $U =\{x_1, x_2, \ldots , x_m\}$, $A=\{a_1, a_2, \ldots , a_n\}$, and $I \subseteq U\times A$ is a binary relation between $U$ and  $A$. Here each $x_i(i\leq m)$ is called an object, and each $a_j(j \leq n)$ is called an attribute.  $xIa$ or $(x, a) \in I$ indicates that an object $x\in U$ has the
attribute $a\in A$.

\textbf{Definition 2}. \cite{Ganter} Let $(U, A, I)$ be a formal context. For any $X \subseteq U$ and $B \subseteq A$, two operations are defined respectively:

$^* : P(U)\rightarrow P(A)$, $X^* = \{m \in A|\forall g \in X, (g,m) \in I\}$

$^* : P(A)\rightarrow P(U)$, $B^* = \{g \in U | \forall m \in B, (g,m) \in I\}$

The following are the definitions of formal concepts and concept lattices.

\textbf{Definition 3}. \cite{Ganter} Let $(U, A, I)$ be a formal context. For any $X \subseteq U$ and $B \subseteq A$, if $X^*=B$ and $B^*=X$, then the pair $(X, B)$   is called a formal concept, where $X$ and $B$
are called the extension and the intension of $(X, B)$,
respectively.
For concepts $(X_1, B_1), (X_2, B_2)$, where $X_1, X_2 \subseteq U$, $ B_1, B_2 \subseteq A$, one can define the partial order as follows:

$(X_1, B_1) \leq (X_2, B_2) \Leftrightarrow X_1 \subseteq X_2\Leftrightarrow B_2 \subseteq B_1$

Furthermore, we have the following definitions:

$(X_1, B_1) \bigwedge (X_2, B_2) = (X_1 \cap X_2, (B_1 \cup B_2)^{**})$ or  $(X_1 \cap X_2, (X_1 \cap X_2)^{*})$

$(X_1, B_1) \bigvee (X_2, B_2) = ((X_1 \cup X_2)^{**}, B_1 \cap B_2)$ or  $((B_1 \cap B_2)^{*}, B_1 \cap B_2)$

Thus, all formal concepts from $(U, A, I)$ form a complete lattice, and it is defined as a concept lattice and denoted by $L(U, A, I)$.

\textbf{ Lemma 1}. \cite{Ganter}
 For any $X_1, X_2, X \subseteq U$, $ B_1, B_2, B \subseteq A$, here $(U, A, I)$ is a formal context, the following statements hold:

(1) $X_1\subseteq X_2 \Rightarrow X^*_2 \subseteq X^*_1$, $\ B_1\subseteq B_2 \Rightarrow B^*_2\subseteq B^*_1$;

(2) $X \subseteq X^{**}$, $\ B \subseteq B^{**}$;

(3) $X^* = X^{***}$, $\ B^* = B^{***}$;

(4) $X \subseteq B^{*}\ \Leftrightarrow\ B \subseteq X^{*}$;

(5) $(X_1\cup X_2)^* = X^*_1\cap X^*_2, \ (B_1\cup B_2)^* = B^*_1\cap B^*_2$;

(6) $(X_1\cap X_2)^* \supseteq   X^*_1\cup X^*_2, \ (B_1\cap B_2)^* \supseteq B^*_1\cup B^*_2$.

\

 A formal context is typically represented by a  table of $0$ and $1$, with $1$s meaning binary relations between objects (rows) and attributes (columns). A simple example of a formal context is presented as follows:

 \begin{table*}[ht]
            \begin{center}
                \begin{normalsize}
                    \caption{Formal context $(G, M, I)$}
                    \label{table_tab1}
                    \begin{tabular}
                        {|c|c c  c  c  c|}
                        \hline
                        $G$     & $\ \ \ \ a_1\ \ \ \ \ $       &  $\ \ \ \ a_2\ \ \ \ \ $  &  $\ \ \ \ a_3\ \ \ \ \ $  & $\ \ \ \ a_4\ \ \ \ \ $  &  $\ \ a_5$\ \ \  \\
                       \hline
 1 & 0   & 1  &  1  & 0 &  0  \\

 2 & 1  &  1  & 0 &  0  &  0   \\

 3 & 1   & 0 &  0  &  0  &  0   \\

4 & 0 &  0  &  0  &  0  &   1  \\

5 & 0 &  0  &  0  &  1 &   1  \\

6 & 0 &  0  &  1  &  1 &   1  \\

7 & 1 &  1  &  1  &  0 &   0  \\
\hline
                    \end{tabular}
                \end{normalsize}
            \end{center}
        \end{table*}

 The formal concepts in  Table \ref{table_tab1}  can be calculated as  given in the following Table \ref{table_2tab}:

 \begin{table*}[ht]
            \begin{center}
                \begin{scriptsize}
                \caption{Formal concepts in  Table 1}
                    \label{table_2tab}
                    \begin{tabular}
                        {c c c c  }
                        \hline
                             &        &  $C_0=(\{1,2,3,4,5,6,7\},\emptyset)$  &       \ \ \  \\

                   &    &     &        \\

 $C_4=(\{2,3,7\},\{a_1\}) $ & $C_3=(\{1,2,7\},\{a_2\}) $   & $C_2=(\{1,6,7\},\{a_3\}) $  &  $C_1=(\{4,5,6\},\{a_5\}) $     \\

   &    &     &         \\

  & $C_5=(\{2,7\},\{a_1, a_2\}) $  &  $C_6=(\{1,7\},\{a_2, a_3\}) $   & $C_7=(\{5,6\},\{a_4, a_5\}) $      \\

  &    &     &         \\

  &  $C_9=(\{7\},\{a_1, a_2, a_3\}) $ &    &  $C_8=(\{6\},\{a_3, a_4, a_5\}) $    \\

   &    &     &         \\

 &  &    $C_{10}=(\emptyset,\{a_1, a_2, a_3 ,a_4, a_5\}) $   &      \\
\hline

                    \end{tabular}
                \end{scriptsize}
            \end{center}
        \end{table*}

  In fact, one can visualise a formal concept in a  table of $0$ and $1$  as a closed rectangle of $1$s, where the rows and columns  are not necessarily contiguous \cite{Andrews}.
  Suppose we define the cell of the $i$th row and $j$th column as $(i,j)$. Thus in Table \ref{table_tab1}, $(5,4)$, $(5,5)$, $(6,4)$ and $(6,5)$ form the concept $C_7$, and $C_7$ is a rectangle of height $2$ and width $2$. Similarly $(6,3)$, $(6,4)$ and $(6,5)$ form the concept $C_8$, and $C_8$ is a rectangle of height $1$ and width $3$.  $(1,3)$, $(6,3)$ and $(7,3)$ form the concept $C_2$, and $C_2$ is a rectangle of height $3$ and width $1$, here $(1,3)$ and $(6,3)$ are not contiguous.

On the other hand, a formal concept can be obtained by applying the $^*$ operator to a set of attributes to get its extension, and then applying the $^*$ operator to the extension to get the intension.
From the context in  Table \ref{table_tab1},  $\{a_4\}^*= \{5,6\}$ and $\{5,6\}^*=\{a_4,a_5\}$. So $(\{5,6\}, \{a_4,a_5\})$ is concept $C_7$ in  Table \ref{table_2tab}.
From the rectangle theory of a formal concept, $C_7$ is the biggest rectangle to cover $(5,4)$, $(5,5)$, $(6,4)$ and $(6,5)$ simultaneously.

Based on the rectangle theory of a formal concept, next we will discuss the concept reduction preserving binary relations.

\section{Concept reduction preserving binary relations} \label{s2}

We first present the definition of concept reduction preserving binary relations as follows.

\textbf{Definition 4}. \cite{Cao,Wei} Let $(G, M, I)$ be a formal context. $\mathcal{F}\subseteq L(G, M, I)$ is called a  consistent concept set preserving binary relations
if

\[ I=\bigcup\limits_{(A_i,B_i)\in \mathcal{F}} A_i\times B_i   \]

Further, $\mathcal{F}\subseteq L(G, M, I)$ is called a  concept reduction set preserving binary relations if for any $(A,B)\in \mathcal{F}$

\[ I\neq \bigcup\limits_{(A_i,B_i)\in \mathcal{F}\backslash \{(A,B) \}} A_i\times B_i   \]

It follows from  Definitions 4 that there
exists at least one concept reduction set preserving binary relations from $L(G, M, I)$ \cite{Cao}.

It is obvious that \[ \mathcal{O}(G,M,I)=\{ (g^{**},g^*)  | g\in G \} \] is a  consistent concept set preserving binary relations.
 \[ \mathcal{A}(G,M,I)=\{ (m^{*},m^{**})  | m\in M \} \] is also a  consistent concept set preserving binary relations \cite{Cao}.

For example, in Table \ref{table_tab1}, $\{c_1, c_2, c_3, c_4, c_7 \}$ is a  concept reduction set preserving binary relations for formal concepts in  Table 1. In other words, $\{c_1, c_2, c_3, c_4, c_7 \}$ covers exactly binary relations  in  Table 1.
 $\{c_1, c_4, c_5, c_6, c_7,  c_8\}$ is also a  concept reduction set preserving binary relations, thus the concept reduction set is not unique.

\

Based on Definition 4, the significance of a concept in a given  formal context
may be different and they play different roles in
the concept reduction. These concepts can be
classified into three types: the core
concepts, the relatively necessary
concepts, and the  unnecessary
concepts. Next we provide the characteristics
of these types of concepts.

\textbf{Definition 5}. \cite{Cao,Wei} Let $(G, M, I)$ be a formal context.
$\mathcal{F}_i\subseteq L(G, M, I)$, where $ i\in \tau$, are all
concept reductions of $L(G, M, I)$, then the concepts in $L(G, M, I)$ are classified
into three types:

(1) Core concept set:
$L =\bigcap\limits_{i\in \tau}\mathcal{F}_i$

(2) Relatively necessary concept set:$ M = \bigcup\limits_{i\in \tau}\mathcal{F}_i-\bigcap\limits_{i\in \tau}\mathcal{F}_i$

(3) Unnecessary  concept set: $N =L(G, M, I)-\bigcup\limits_{i\in \tau}\mathcal{F}_i$

For example,  in Table \ref{table_tab1}, $(5,4)$ is only covered by concept $c_7$, so  $c_7$
  is a core concept. It is easy to show that $c_1$ and $c_4$
  are also  core concepts. Thus, $c_2$,  $c_3$,  $c_5$,  $c_6$ and $c_8$ are relatively necessary concepts.
  Later we will explain why $c_9$ is an unnecessary concept.

Based on the rectangle theory of formal concept, the following judgment theorem of the core concept is given.

\textbf{ Theorem 1}. \cite{Cao,Wei} Let $(G, M, I)$ be a formal context. For concept $(A,B)\in L(G, M, I)$, if $(g,m)\in(A,B)$ is only covered by $(A,B)$, then concept $(A,B)$ is a core concept.

In what follows, our focus   is the discussions of  relatively necessary concept and unnecessary concept. First we present the following lemma.

\textbf{ Lemma 2}. Let $(G, M, I)$ be a formal context. For concept $(A,B)\in L(G, M, I)$, if there exists a set $A'\subseteq G\backslash A$
such that $B=\bigcup\limits_{g\in A'}g^*$,
or there exists a set $B'\subseteq M\backslash B$
such that $A=\bigcup\limits_{m\in B'}m^*$, then $(A,B)$ is an  unnecessary concept, and $(A,B)$ is called a  side covered concept.

\begin{proof}  Without
loss of generality, we assume that $B=\bigcup\limits_{g\in A'}g^*$.

For any $(x,y)\in (A,B)$, suppose that $y\in g^*$. As $g\notin A$, so $(g,y)\notin (A,B)$.

For any $(C,D)$, such that $(g,y)\in (C,D)$, by  Lemma 1.(1),
\[g\in C \Rightarrow D\subseteq g^* \Rightarrow D\subseteq B \Rightarrow A\subseteq C  \Rightarrow (x,y)\in (C,D)\]

Suppose that
 $\mathcal{F}\subseteq L(G, M, I)$ is   a  consistent concept set,  then $\mathcal{F}\backslash \{(A,B)\}$ is still  a  consistent concept set.
 Thus $(A,B)$ is an  unnecessary concept.
\end{proof}

For example, in Table \ref{table_tab1}, $\{a_1, a_2, a_3\}=\{1\}^* \cup \{2\}^*$, so $C_9=(\{7\},\{a_1, a_2, a_3\}) $ is an  unnecessary concept.

If we visualise a formal concept as a rectangle, then the following lemma is obvious.

\textbf{ Lemma 3}. Let $(G, M, I)$ be a formal context.  Concept $(A,B)\in L(G, M, I)$ must be unnecessary if $(A,B)$ is covered by some core concepts, where cover means for any $(g,m)\in (A,B)$, $(g,m)$ is also in a core concept.

 \begin{table*}[ht]
            \begin{center}
                \begin{normalsize}
                    \caption{Formal context $(G, M, I_3)$}
                    \label{table_tab03}
                    \begin{tabular}
                        {|c|c c  c  c  c|}
                        \hline
                        $G$     & $\ \ \ \ a_1\ \ \ \ \ $       &  $\ \ \ \ a_2\ \ \ \ \ $  &  $\ \ \ \ a_3\ \ \ \ \ $  & $\ \ \ \ a_4\ \ \ \ \ $  &  $\ \ a_5$\ \ \  \\
                       \hline
 1 & 1   & 1  &  1  & 0 &  0  \\

 2 & 1  &  1  & 1 &  0  &  0   \\

 3 & 1   & 1 &  1  &  0  &  0   \\

4 & 0 &  1  & 1  & 1   &   1  \\

5 & 0 &  1  &  1  &  1 &   1  \\

6 & 0 &  1  &  1  &  1 &   1  \\

\hline
                    \end{tabular}
                \end{normalsize}
            \end{center}
        \end{table*}

For example, in Table \ref{table_tab03},   there are concepts $C_1=(\{4,5,6\},\{ a_2, a_3, a_4, a_5\}) $, $C_2=(\{1,2,3\},\{a_1, a_2, a_3\}) $ and $C_3=(\{1,2,3,4,5,6\},\{ a_2, a_3\}) $.  $(1, a_1)$ is only covered by  $C_2$, so $C_2$ is a core   concept.  $(4, a_5)$ is only covered by  $C_1$, so $C_1$ is a core   concept. If we visualise a formal concept as a rectangle,   $C_3$ is covered by  $C_2$ and  $C_1$, so   $C_3$ is an unnecessary concept.

\

Based on Lemma 2 and Lemma 3, we have the following  judgment theorem of unnecessary concept.

\textbf{ Theorem 2}.   Let $(G, M, I)$ be a formal context. For   $(A,B)\in L(G, M, I)$ is not a core concept,  $(A,B)$ is an unnecessary concept if and only if $(A,B)$ is covered by some core concepts, or covered by some side covered concepts, or covered by both some core concepts, and some side covered concepts, where cover means for any $(g,m)\in (A,B)$, $(g,m)$ is also in a core concept, or in a side covered concept.

\begin{proof}

 $\Leftarrow$ It follows immediately from Lemma 2 and Lemma 3.

$\Rightarrow$
Let
 \[ G' = G\backslash A,  \mathcal{O}'(G,M,I)=\{ (g^{**},g^*)  | g\in G' \} \]
 \[     M' = M\backslash B,  \mathcal{A}'(G,M,I)=\{ (m^{*},m^{**})  | m\in M' \} \]
\[ \mathcal{F} = \mathcal{O}'(G,M,I) \cup  \mathcal{A}'(G,M,I) \]

$(A,B)\in L(G, M, I)$ is not a core concept, thus for each $(g,m)\in (A,B)$, $(g,m)$ is covered by at least two concepts. Therefore, all core concepts are in $\mathcal{F}$.

$(A,B)$ is an unnecessary concept, thus \[ I=\bigcup\limits_{(A_i,B_i)\in \mathcal{F}} A_i\times B_i   \]

Therefore,  $(A,B)$ is covered by some core concepts, or covered by some side covered concepts, or covered by both some core concepts, and some side covered concepts.
\end{proof}

We also have the following  judgment theorem of relatively necessary concept.

\textbf{ Theorem 3}.   Let $(G, M, I)$ be a formal context. For   $(A,B)\in L(G, M, I)$ is not a core concept,  $(A,B)$ is   a relatively necessary concept if and only if \[ I \neq \bigcup\limits_{(A_i,B_i)\in \mathcal{F}} A_i\times B_i   \]
where
\[ G' = G\backslash A,  \mathcal{O}'(G,M,I)=\{ (g^{**},g^*)  | g\in G' \} \]
 \[     M' = M\backslash B,  \mathcal{A}'(G,M,I)=\{ (m^{*},m^{**})  | m\in M' \} \]
\[ \mathcal{F} = \mathcal{O}'(G,M,I) \cup  \mathcal{A}'(G,M,I) \]

\

It should be noted that
with our simple judgment theorems, it is easy to develop a program for computing both  relative necessary concepts and
unnecessary concepts.

 \begin{table*}[ht]
            \begin{center}
                \begin{normalsize}
                    \caption{Formal context $(G, M, I_4)$}
                    \label{table_tab04}
                    \begin{tabular}
                        {|c|c c  c  c  c|}
                        \hline
                        $G$     & $\ \ \ \ a_1\ \ \ \ \ $       &  $\ \ \ \ a_2\ \ \ \ \ $  &  $\ \ \ \ a_3\ \ \ \ \ $  & $\ \ \ \ a_4\ \ \ \ \ $  &  $\ \ a_5$\ \ \  \\
                       \hline
 1 & 1   & 1  &  1  & 0 &  0  \\

 2 & 1  &  1  & 1 &  0  &  0   \\

 3 & 1   & 1 &  1  &  0  &  0   \\

4 & 0 &  1  & 1  & 1   &   1  \\

5 & 0 &  1  &  1  &  1 &   1  \\

6 & 0 &  1  &  1  &  1 &   1  \\

7 & 1 &  1  &  0 &  0 &  0  \\

8 & 1 &  0  &  1 &  0 &  0  \\

\hline
                    \end{tabular}
                \end{normalsize}
            \end{center}
        \end{table*}

For example, in Table \ref{table_tab04},   there are concepts $C_1=(\{4,5,6\},\{ a_2, a_3, a_4, a_5\}) $, $C_2=(\{1,2,3\},\{a_1, a_2, a_3\}) $ and $C_3=(\{1,2,3,4,5,6\},\{ a_2, a_3\}) $.
$\{a_1, a_2, a_3\}=\{7\}^* \cup \{8\}^*$, so $C_2=(\{1,2,3\},\{a_1, a_2, a_3\}) $ is a side covered concept.
  $(4, a_5)$ is only covered by  $C_1$, so $C_1$ is a core   concept. If we visualise a formal concept as a rectangle,   $C_3$ is covered by  $C_2$ and  $C_1$, so   $C_3$ is an unnecessary concept.

In Table \ref{table_tab04},   consider concept $C_4=(\{1,2,3,7\},\{ a_1, a_2\}) $. It is easy to show that  $C_4$ is not a core   concept.
Further,
$ \mathcal{O}'(G,M,I)=\{ (g^{**},g^*)  | g\in G' \} =\{ c_1, c_5\}$, where $C_1=(\{4,5,6\},\{ a_2, a_3, a_4, a_5\}) $, $C_5=(\{1,2,3,8\},\{ a_1, a_3\}) $.
$  \mathcal{A}'(G,M,I)=\{ (m^{*},m^{**})  | m\in M' \} =\{ c_1, c_6\}$,  where $C_6=(\{1,2,3,4,5,6,8\},\{  a_3\}) $. As \[  \mathcal{F}=  \mathcal{O}'(G,M,I) \cup  \mathcal{A}'(G,M,I) =\{ c_1, c_5, c_6\}\] $C_4$ is not covered by $  \mathcal{F}$,
 thus   $C_4$ is a   relatively necessary concept.

\

To further discuss the characteristics of concepts of each type, we consider the maximum number of concepts of each type. As a core concept is in $\mathcal{O}(G,M,I) \cap  \mathcal{A}(G,M,I)$, the following lemma is straightforward.

\textbf{ Lemma 4}. Let $(G, M, I)$ be a formal context, and $m= |G|$, $n=|M|$.  Then the maximum number of core concepts is not greater than
$\min(m,n)$.

 \begin{table*}[ht]
            \begin{center}
                \begin{normalsize}
                    \caption{Formal context $(G, M, I_5)$}
                    \label{table_tab05}
                    \begin{tabular}
                        {|c|c c  c  c  c|}
                        \hline
                        $G$     & $\ \ \ \ a_1\ \ \ \ \ $       &  $\ \ \ \ a_2\ \ \ \ \ $  &  $\ \ \ \ a_3\ \ \ \ \ $  & $\ \ \ \ a_4\ \ \ \ \ $  &  $\ \ a_5$\ \ \  \\
                       \hline
 1 & 1   & 1  &  0  & 0 &  0  \\

 2 & 1  &  0  & 1   &  0  &  0   \\

 3 & 1   & 0 &  0  &  1  &  0   \\

4 & 1 &  0  & 0    & 0   &   1  \\

5 & 0 &  1  &  1   &  0 &   0  \\

6 & 0 &  1  &   0  &  1 &   0  \\

7 & 0 &  1  &  0   &  0 &  1  \\

8 & 0 &  0  &  1 &  1 &  0  \\

9 & 0 &  0  &  1 &  0 &  1  \\

10 & 0 &  0  &  0 &  1 & 1  \\

\hline
                    \end{tabular}
                \end{normalsize}
            \end{center}
        \end{table*}

For example, in Table \ref{table_tab05},  there are 5 concepts in $\mathcal{A}(G,M,I)$, and there are 10 concepts in $\mathcal{O}(G,M,I)$, but there is neither core concepts nor unnecessary concepts. The number of relatively necessary concepts is \[ 10+5= \binom{5 }{2}+5 = \binom{|M| }{\lfloor|M|/2\rfloor}+|M|
  \]

Inspired by this example, we have the following lemma.

\textbf{ Lemma 5}. Let $(G, M, I)$ be a formal context, and   $n=|M|$. Suppose $|G|>\binom{n }{\lfloor n/2\rfloor}$, then    the maximum number of  relatively necessary concepts is not less than \[
\binom{n }{\lfloor n/2\rfloor}+n
  \]
 where $\lfloor x\rfloor$ is the largest number that is less than or equal to $x$.

\begin{proof}

Note that $\binom{n }{k} = \binom{n }{n-k}$, and   $\binom{n }{\lfloor n/2\rfloor}$ is the maximum, where $1\leq k \leq n$.

For the formal context $(G, M, I)$, let each object (corresponding to a row in the table) have $ \lfloor n/2\rfloor $ attributes.
According to combinatorial mathematics, there are $\binom{n }{\lfloor n/2\rfloor}$ different objects. The $k$th  object forms a concept
$(\{k\}^{**}, \{k\}^*)$, where $\{k\}^{**} = \{k\}$, which is not covered by \[ \mathcal{F} = \mathcal{O}'(G,M,I) \cup  \mathcal{A}'(G,M,I) \]
where
\[   \mathcal{O}'(G,M,I)=\{ (g^{**},g^*)  | g\in G\backslash\{k\} \} \]
 \[     \mathcal{A}'(G,M,I)=\{ (m^{*},m^{**})  | m\in M\backslash \{k\}^*\} \]
Thus $(\{k\}^{**}, \{k\}^*)$ of the $k$th  object is relatively necessary.

In the table of formal context $(G, M, I)$, if there exist two same columns, which means the values of the two positions are always the same.
This is a contradiction by  combinatorial mathematics, thus there are $n$ different columns. Similarly, one can prove that
  $(\{j\}^{*}, \{j\}^{**})$,  where $\{j\}^{**}=\{j\}$ and $1\leq j \leq n$,   of the $j$th  attribute is relatively necessary.

  Therefore, there are $\binom{n }{\lfloor n/2\rfloor}+n$ relatively necessary concepts.

\end{proof}

 \begin{table*}[ht]
            \begin{center}
                \begin{normalsize}
                    \caption{Formal context $(G, M, I_6)$}
                    \label{table_tab06}
                    \begin{tabular}
                        {|c|c c  c  c  |}
                        \hline
                        $G$     & $\ \ \ \ a_1\ \ \ \ \ $       &  $\ \ \ \ a_2\ \ \ \ \ $  &  $\ \ \ \ a_3\ \ \ \ \ $  & $\ \  \ \ \ a_4\ \ \ \ $  \   \\
                       \hline
 1 & 1   & 1  &  0  & 0  \\

 2 & 1  &  0  & 1   &  0     \\

 3 & 1   & 0 &  0  &  1      \\

4 & 0 &  1  & 1    & 0      \\

5 & 0 &  1  &  0   &  1    \\

6 & 0 &  0  &   1  &  1    \\

7 & 1 &  1  &  1   &  0    \\

8 & 1 &  1  &  0 &  1    \\

9 & 1 &  0  &  1 &  1   \\

10 & 0 &  1  &  1 &  1    \\

\hline
                    \end{tabular}
                \end{normalsize}
            \end{center}
        \end{table*}

 In Table \ref{table_tab06}, let $C_1=(\{1,7 \},\{a_1, a_2 \}) $,  $C_2=(\{2,7 \},\{a_1, a_3 \}) $, $C_7=(\{7 \},\{a_1, a_2, a_3 \})$, then
 $C_7$ is covered by $C_1$ and $C_2$, thus $C_7$ is an unnecessary concept. In fact,
 if we add one more object with 3 attributes, then this object (or row) must form an unnecessary concept. Thus we have the following lemma.

\textbf{ Lemma 6}. Let $(G, M, I)$ be a formal context, and   $n=|M|$. Suppose $|G|>\binom{n }{\lfloor n/2\rfloor} + \binom{n }{\lfloor n/2 + 1\rfloor}$, then    the maximum number of   unnecessary concepts is not less than \[
\binom{n }{\lfloor n/2 + 1\rfloor}
  \]
 where $\lfloor x\rfloor$ is the largest number that is less than or equal to $x$.

\

From the discussion above, we know that both relatively necessary concepts and unnecessary concepts are exponential time problems.

\
%
%Let $(G, M, I)$ be a formal context, then $\mathcal{A}(G,M,I)=\{ (m^{*},m^{**})  | m\in M \}$ is a consistent concept set preserving binary relations. After deleting all unnecessary concepts in $\mathcal{A}(G,M,I)$, then we get a concept reduction.
%Based on this idea, we have the following algorithm for finding concept reduction.

\section{An algorithm for attribute reduction\label{s03}}

Based on the discernibility matrix and
Boolean function, a method of attribute reduction is proposed while preserving the extensions
in concept lattices in \cite{Zhang, Zhang05}.  Similarly,
 two kinds of attribute reduction are proposed in the decision formal context based on the maximal rules \cite{Li}, and an approach to computing attribute reductions is proposed based on the discernibility matrix and Boolean function. However, the approaches in \cite{Zhang, Zhang05,Li} are very difficult to implemented.

Note that the attribute reduction in both \cite{Zhang, Zhang05} and \cite{Li}  can be simplified to  reducing attributes while preserving the extensions for a set of formal concepts. We also know that  for any formal concept $(X,B)$ of $L(U, A, I)$,   $X^*=B$ and $B^*=X$.
Based on this explicit idea, a fast algorithm for  reducing attributes while preserving the extensions for a set of  formal concepts is proposed in this paper.
Technically, by checking   $X^{**}=X$ to determine whether to delete a column of formal context or not.

As the new algorithm would store both formal context and extension of a concept as a  bit-array, and we can use bit-operations to process set operations "or" as well as "and". What is more, there is no need to consider other formal concepts in $L(U, A, I)$, thus the algorithm, given below with analysis, is very fast.

    %[h]当前位置。将图形放置在正文文本中给出该图形环境的地方。如果本页所剩的页面不够，这一参数将不起作用。
    \begin{figure}[htp] \centering
        \includegraphics[scale=0.6]{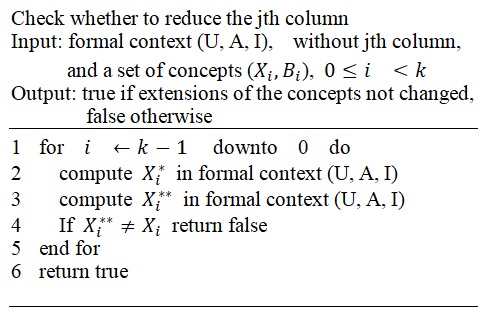}
        \label{figInClose2}
    \end{figure}

\

Line 2 -- Construct an intension of $X$, which could be different from $B$.
It is implemented in C language as the following.

		memset(Row, 0, sizeof(Row));	// clear the Row, which is used to store the intension

		for(q=0; q\textless n; q++) $\{$

            \ \ \ \ 	for(int k = (m - 1)/32; k \textgreater= 0; k- -)	$\{$

             \ \ \ \  \ \ \ \ \ \ \ \ // check whether the qth column contains the ith  extension column

			 \ \ \ \  \ \ \ \ \ \ \ \ if((contextCol[k][q] \& extensionCol[k][i]) != extensionCol[k][i])break;

	        \ \ \ \ $\}$

            \ \ \ \  // the qth column contains the ith  extension column, so to change the intension

			   \ \ \ \ if(k\textless 0)	Row[q\textgreater\textgreater 5] $|=$ (1\textless\textless(q\%32));

		$\}$

Note that here  $m=|U|, n=|A|$, and both formal context contextCol  and extension of a concept extensionCol  are stored as bit-arrays (in the form of 32-bit unsigned integers).  Without processing each object in the formal context and extension of a concept, the bit and operation is used when check whether the $q$th column of  formal context contains the $i$th  extension column.

First call the procedure to check whether to reduce the first column. If return true, then let the  first column of the formal context
 $(U, A, I)$ be zero, otherwise there is no change. Second call the procedure to check whether to reduce the  second  column. If return true, then let the  second column of the formal context
 $(U, A, I)$ be zero, otherwise there is no change. Keep this process going and at the end of this process, we get an attribute reduction  preserving the extensions for a set of  formal concepts.

 Suppose that first call the procedure to check whether to reduce the second column, second call the procedure to check whether to reduce the third  column.  Finally call the procedure to check whether to reduce the first column. At the end of this process, we may get a different  attribute reduction  preserving the same set of  formal concepts.

 Thus by calling the procedure with different starting $i$th column, where $1\leq i\leq n$, we can get all attribute reductions  preserving the same set of  formal concepts.

 \

Some experiments are done  to compute attribute reductions with the new  algorithm.
The   experiment results are given in Table \ref{table_tab05}. Here mushroom data and nursery data are from \cite{Frank}.
The experiments are carried out using a laptop computer with an Intel Core i5-2450M 2.50 GHz processor and 8GB of RAM.

 \begin{table*}[ht]
            \begin{center}
                \begin{normalsize}
                    \caption{Computation of attribute reductions with the new  algorithm}
                    \label{table_tab05}
                    \begin{tabular}
                        { c c c  c  c  }
                        \hline
                            & Mushroom       &  Nursery        \ \  \\
                       \hline
$|G|\times |M|$ & $8,124\times 115$   & $12,960\times 30$  &        \\

 Number of concepts  & 512               &  512                       \\

 Time in seconds & 4.218                  &  0.582                            \\

 Number of reduced columns & 19                 &   2                        \\
\hline
                    \end{tabular}
                \end{normalsize}
            \end{center}
        \end{table*}

In Table \ref{table_tab05}, In-Close4 algorithm from \cite{Andrews17} is used to compute all the formal concepts for mushroom data. From which, we take the first 512  concepts.  With our new  algorithm, 19 columns of formal context can be reduced while preserving the extensions for 512 formal concepts. Experiments demonstrate that the new algorithm is much  effective in the computation of attribute reductions.

\newpage

\

\section{Conclusions and future work\label{s5}}

 A pair $(X, B)$   is called a formal concept if $X^*=B$ and $B^*=X$, which is a frequently used definition. With another perspective,
based on rectangle theory of formal concept and set covering theory, the concept reduction preserving binary
relations was discussed. It is known that there are three types of formal concepts:  core concepts, relative necessary concepts and
unnecessary concepts. We presented the new judgment theorems for relative necessary concepts and
unnecessary concepts. Also, we  gave  bounds for both the maximum number of relative necessary concepts and the maximum number of unnecessary concepts. From our results, it is known that both relatively necessary concepts and unnecessary concepts are exponential time problems.

As the attribute reduction approaches in both \cite{Zhang} and \cite{Li}  can be simplified to  reducing attributes while preserving the extensions for a set of formal concepts,
a fast algorithm was proposed for  reducing attributes while preserving the extensions for a set of  formal concepts. Experiments demonstrated that the new algorithm is much  effective  in the computation of attribute reductions.

Object oriented concept lattice is a more extensive concept lattice \cite{Yao}. So, it is more difficult to reduce concepts or attributes in object oriented concept lattices.
In the future, we will apply the data structure and technique in these algorithms  to  object oriented concept lattices, %(what is the objected oriented concept? do not talk a direction randomly) ,
attribute oriented concept lattices and so on.

%\noindent\textbf{Acknowledgments} \\
%
%\noindent The authors are grateful to  the referee for useful
%suggestions.

\end{document}